\documentclass{article}
\usepackage{ijcai21}

\usepackage{times}
\usepackage{soul}
\usepackage{url}
\usepackage[hidelinks]{hyperref}
\usepackage[utf8]{inputenc}
\usepackage[small]{caption}
\usepackage{graphicx}
\usepackage{amsmath}
\usepackage{amsthm}
\usepackage{amsfonts}
\usepackage{booktabs}
\usepackage{algorithmic}
\usepackage{natbib}

\usepackage[linesnumbered,lined,ruled]{algorithm2e}

\usepackage{xcolor,colortbl}
\definecolor{mygray}{gray}{0.9}

\newcommand{\eref}[1]{Eq. (\ref{#1})}

\usepackage{caption}
\usepackage{subcaption}

\title{Physics-informed Spline Learning for Nonlinear Dynamics Discovery}

\author{
Fangzheng Sun$^1$\and
Yang Liu$^{2}$\footnote{Corresponding Author}\and
Hao Sun$^{1,3}$\\
\affiliations
$^1$ Department of Civil and Environmental Engineering, Northeastern University, Boston, MA, USA\\
$^2$ Department of Mechanical and Industrial Engineering, Northeastern University, Boston, MA, USA\\
$^3$ Department of Civil and Environmental Engineering, MIT, Cambridge, MA, USA\\

\emails
\{sun.fa, yang1.liu, h.sun\}@northeastern.edu
}

\begin{document}

\maketitle

\begin{abstract}
    Dynamical systems are typically governed by a set of linear/nonlinear differential equations. Distilling the analytical form of these equations from very limited data remains intractable in many disciplines such as physics, biology, climate science, engineering and social science. To address this fundamental challenge, we propose a novel Physics-informed Spline Learning (PiSL) framework to discover parsimonious governing equations for nonlinear dynamics, based on sparsely sampled noisy data. The key concept is to (1) leverage splines to interpolate locally the dynamics, perform analytical differentiation and build the library of candidate terms, (2) employ sparse representation of the governing equations, and (3) use the physics residual in turn to inform the spline learning. The synergy between splines and discovered underlying physics leads to the robust capacity of dealing with high-level data scarcity and noise. A hybrid sparsity-promoting alternating direction optimization strategy is developed for systematically pruning the sparse coefficients that form the structure and explicit expression of the governing equations. The efficacy and superiority of the proposed method have been demonstrated by multiple well-known nonlinear dynamical systems, in comparison with two state-of-the-art methods.
\end{abstract}

\section{Introduction}

Nonlinear dynamics is commonly seen in nature (in many disciplines such as physics, biology, climate science, engineering and and social science), whose behavior can be analytically described by a set of nonlinear governing differential equations, expressed as
\begin{equation}\label{dynamics}
    \dot{\mathbf{y}}(t) = \mathcal{F}(\mathbf{y}(t))
\end{equation}
where $\mathbf{y}(t) = \{y_1(t), y_2(t), ..., y_n(t)\}\in \mathbb{R}^{1\times n}$ denotes the system state at time $t$, $\mathcal{F}(\cdot)$ a nonlinear functional defining the equations of motion and $n$ the system dimension. Note that $\dot{\mathbf{y}}(t) = d\mathbf{y}/dt$. There remain many underexplored dynamical systems whose governing equations (e.g., the exact or explicit form of $\mathcal{F}$) or physical laws are unknown. For example, the mathematical description of an uncharted biological evolution process might be unclear, which is in critical need for discovery given observational data. Nevertheless, distilling the analytical form of the equations from \textit{scarce and noisy data}, commonly seen in practice, is an intractable challenge.

Data-driven discovery of dynamical systems dated back decades \citep{dvzeroski1993discovering,dzeroski1995discovering}. Recent advances in machine learning and data science encourage attempts to develop methods to uncover equations that best describe the underlying governing physical laws. One popular solution relies on symbolic regression with genetic programming \citep{billard2003statistics}, which has been successfully used to distill mathematical formulas (e.g., natural/control laws) that fit data \citep{bongard2007automated,schmidt2009distilling,quade2016prediction,kubalik2019symbolic}. However, this type of approach does not scale well with the system dimension and generally suffers from extensive computational burden when the dimension is high that results in a large search space. Another progress leverages symbolic neural networks to uncover analytical expressions to interpret data \citep{martius2016extrapolation,sahoo2018learning,kim2019integration,long2019pde}, where commonly seen mathematical operators are employed as symbolic activation functions to establish intricate formulas through weight pruning. Nevertheless, this existing framework is primarily built on empirical pruning of the weights, thus exhibits sensitivity to user-defined thresholds and may fall short to produce parsimonious equations for complex systems. Moreover, this method requires numerical differentiation of measured system response to feed the network for discovery of dynamics in the form of \eref{dynamics}, leading to substantial inaccuracy especially when the measurement data is sparsely sampled with large noise.

Another alternative approach reconstructs the underlying equations based on a large-space library of candidate terms and eventually turns the discovery problem to sparse regression \citep{wang2016data,brunton2016discovering}.In particular, the breakthrough work by \citet{brunton2016discovering} introduced a novel paradigm called Sparse Identification of Nonlinear Dynamics (SINDy) for data-driven discovery of governing equations, based on a sequential threshold ridge regression (STRidge) algorithm which recursively determines the sparse solution subjected to pre-defined or adaptive hard thresholds \citep{brunton2016discovering,rudy2017data,champion2019data}. This method has drawn tremendous attention in recent years, showing successful applications in biological systems \citep{mangan2016inferring}, fluid flows \citep{loiseau2018sparse}, predictive control \citep{kaiser2018sparse}, continuous systems (described by partial differential equations (PDEs)) \citep{rudy2017data,schaeffer2017learning,zhang_ma_2020}, etc. Compared with the above symbolic models, SINDy is less computationally demanding thus being more efficient. Nonetheless, since this method relies heavily on the numerical differentiation as target for derivative fitting, it is very sensitive to both data noise and scarcity. Another limitation is that SINDy is unable to handle non-uniformly sampled data while data missing is a common issue in practical applications. 

One way to tackle these issues is to build a differentiable surrogate model to approximate the system state which best fits the measurement data meanwhile satisfying the governing equations to be discovered (e.g., by SINDy). Several recent studies \citep{berg2019data,chen2020deep,Both2019} have shown that deep neural networks (DNNs) can serve as the approximator where automatic differentiation is used to calculate essential derivatives required for reconstructing the governing PDEs. Nevertheless, since DNNs are rooted in global universal approximation, it is extremely computationally demanding to obtain a fine-tuned model with high accuracy of local approximation (crucial for equation discovery), especially when the nonlinear dynamics is very complex, e.g., chaotic. To overcome this issue, we take advantage of cubic B-splines, a powerful local approximator for time series. Specifically, we develop a Physics Informed Spline Learning (PiSL) approach to discover sparsely represented governing equations for nonlinear dynamics, based on \textit{scarce and noisy data}. The key concept is to (1) leverage splines to interpolate locally the dynamics, perform analytical differentiation and build the library of candidate terms, (2) reconstruct the governing equations via sparse regression, and (3) use the equation residual as constraint in turn to inform the spline approximation. The synergy between splines and discovered equations leads to the robust capacity of dealing with high-level data scarcity and noise. A hybrid sparsity-promoting alternating direction optimization strategy is developed for systematically training the spline parameters and pruning the sparse coefficients that form the structure and explicit expression of the governing equations. The efficacy of PiSL is finally demonstrated by two chaotic systems under different conditions of data volume and noise.

\section{Methodology}\label{Method}
In this section, we state and explain the concept and algorithm of PiSL for discovering governing equations for nonlinear dynamics, including introduction to cubic B-Splines, the basic network architecture, and physics-informed network training (in particular, a hybrid sparsity-promoting alternating direction optimization approach).

\begin{figure*}[t!]
	\centering
	\includegraphics[width=0.81\linewidth]{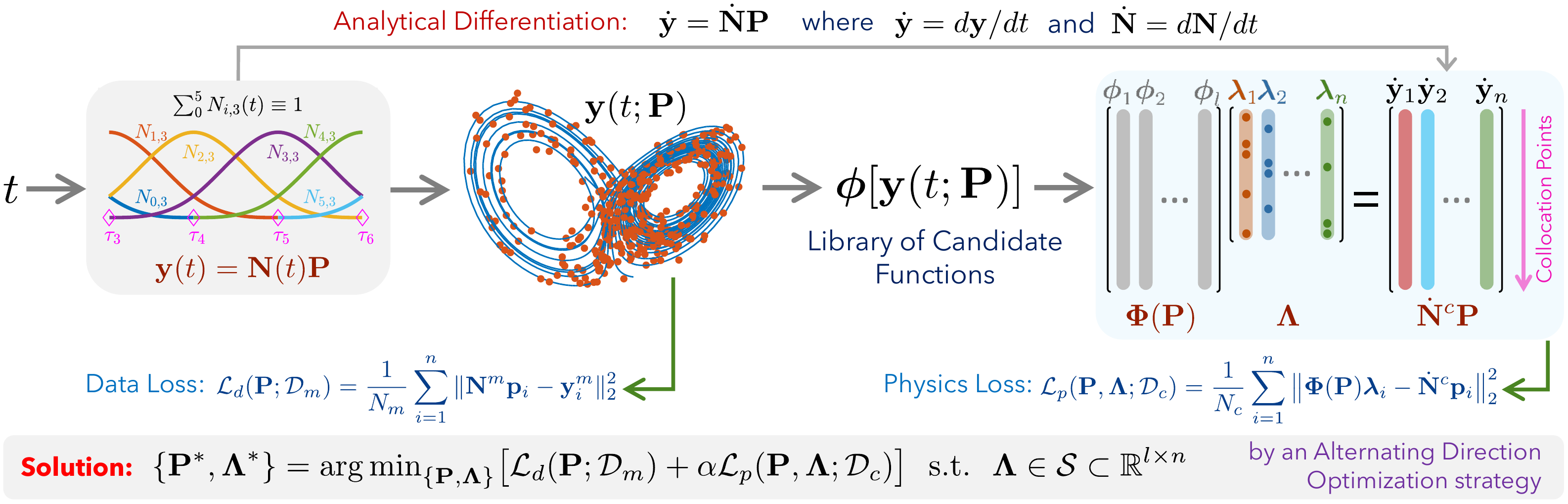} 
	\caption{Schematic architecture of PiSL for discovery of governing equations for nonlinear dynamics based on scarce and noisy data.} \vspace{-6pt}
	\label{PiSL}
\end{figure*}

\subsection{Cubic B-Splines}\label{spline learning}

Cubic B-splines, as one type of piece-wise polynomials of degree 3 with continuity at the knots between adjacent segments, have been widely used for curve-fitting and numerical differentiation of experimental data. The basis function $\mathbf{N}(t)$ between two knots are defined as:
\begin{equation} \label{basis}
\resizebox{.91\linewidth}{!}{$
    \displaystyle
\begin{split}
    N_{s,0}(t) & =
    \begin{cases}
      1 & \text{if $\tau_s \leq t < \tau_{s+1}$} \\
      0 & \text{otherwise}
    \end{cases} \\
    N_{s,k}(t) & = \dfrac{t-\tau_s}{\tau_{s+k}-\tau_i}N_{s,k-1}(t)+
                   \dfrac{\tau_{s+k+1}-t}{\tau_{s+k+1}-\tau_{s+1}}N_{s+1,k-1}(t)
\end{split}
$}
\end{equation}
where $\tau_s$ denotes knot, $k$ denotes degree of polynomial (e.g., $k=3$), and $t$ can be any point in the given domain. Cubic B-spline interpolation  is calculated by multiplying the values of the nonzero basis functions with a set of equally spaced control points $\mathbf{p}\in\mathbb{R}^{(r+3)\times1}$, namely, $y(t)=\sum_{s=0}^{r+2}N_{s,3}(t)p_s$, where the number of control points $r+3$ is chosen empirically, mainly in accordance with the frequency of system state while accounting for computational efficiency, i.e., small number for smooth response to avoid overfitting and large number for intense oscillation to fully absorb measurement information. The optimal set of control points will produce the splines that best fit the given information (e.g., data). In the case of approximating nonlinear dynamics by cubic B-splines, we place $r+7$ knots to cover the calculation of lower degree basis functions, in a non-decreasing order in time domain $[0, T]$, denoted by $\tau_{0},\tau_{1},\tau_{2},... ,\tau_{r+6}$ where $\tau_{0}<\tau_{1}<\tau_{2}<\tau_{3}=0$ and $\tau_{r+3}=T<\tau_{r+4}<\tau_{r+5}<\tau_{r+6}$. One important feature of cubic B-splines is that the functions are analytically differentiable (e.g., $\dot{\mathbf{N}}$ can be obtained based on \eref{basis}) and and their first and second derivatives are all continuous, which allows us to fit not only data but also the differential equations shown in \eref{dynamics}.

\subsection{Network Architecture}
We start with an overview of the PiSL network architecture, as depicted in Figure \ref{PiSL}. We first define $n$ sets of control points for cubic B-splines $\mathbf{P}=\{\mathbf{p}_1, \mathbf{p}_2, ..., \mathbf{p}_n\}\in\mathbb{R}^{(r+3)\times n}$, which are multiplied with the spline basis $\mathbf{N}(t)$ to interpolate the $n$-dimensional system state:
\begin{equation}\label{spline}
    \mathbf{y}(t; \mathbf{P}) = \mathbf{N}(t)\mathbf{P}
\end{equation}
Thus, we can obtain $\dot{\mathbf{y}}(\mathbf{P}) = \dot{\mathbf{N}}\mathbf{P}$ based on analytical differentiation. We assume that the form of $\mathcal{F}(\cdot)$ in \eref{dynamics} is governed by only a few important terms which can be selected from a library of $l$ candidate functions $\boldsymbol{\phi}(\mathbf{y})\in\mathbb{R}^{1\times l}$ \citep{brunton2016discovering} which consists of many candidate terms, e.g., constant, polynomial, trigonometric and other functions :
\begin{equation}\label{candidate}
    \boldsymbol{\phi} = \left\{ 1, \mathbf{y}, \mathbf{y}^2, ..., \sin(\mathbf{u}), \cos(\mathbf{y}), ..., \mathbf{y}\odot\sin(\mathbf{y}), ... \right\}
\end{equation}
where $\odot$ denotes the element-wise Hadamard product. With the interpolated state variables and their analytical derivatives, the governing equations can be written as:
\begin{equation}\label{library}
    \dot{\mathbf{y}}(\mathbf{P}) = \boldsymbol{\phi}(\mathbf{P})\boldsymbol{\Lambda}
\end{equation}
where $\boldsymbol{\phi}(\mathbf{P}) = \boldsymbol{\phi}(\mathbf{y}(t; \mathbf{P}))$; $\boldsymbol{\Lambda}=\{\boldsymbol{\lambda}_1, \boldsymbol{\lambda}_2,..., \boldsymbol{\lambda}_n\}\in\mathbb{R}^{l \times n}$ is the coefficient matrix belonging to a constraint subset $\boldsymbol{\mathcal{S}}$ satisfying sparsity (only the active candidate terms in $\boldsymbol{\phi}$ have non-zero values), e.g., $\boldsymbol{\Lambda} \in \boldsymbol{\mathcal{S}} \subset \mathbb{R}^{l \times n}$. Hence, the discovery problem can then be stated as: denoting the measurement domain as $m$ and given the measurement data $\mathcal{D}_m=\{\mathbf{y}_i^m\}_{i=1,...,n}\in\mathbb{R}^{N_m\times n}$, find the best set of $\mathbf{P}$ and $\boldsymbol{\Lambda}$ such that \eref{library} holds $\forall t$. Here, $\mathbf{y}_i^m$ is the measured response of the $i$th state and $N_m$ is the number of data points in the measurement. The loss function for training the PiSL network consists of the \textit{data} ($\mathcal{L}_{d}$) and \textit{physics} ($\mathcal{L}_{p}$) components, expressed as follows:  
\begin{equation} \label{loss1}
    \mathcal{L}_{d}(\mathbf{P}; \mathcal{D}_m) = \sum_{i=1}^n\frac{1}{N_m}\left\|\mathbf{N}^{m}\mathbf{p}_i-\mathbf{y}_i^m\right\|_2^2
\end{equation}
\begin{equation} \label{loss2}
    \mathcal{L}_{p}(\mathbf{P}, \boldsymbol{\Lambda}; \mathcal{D}_c) = \sum_{i=1}^n\frac{1}{N_c}\big\|\boldsymbol{\Phi}(\mathbf{P})\boldsymbol{\lambda}_i - \dot{\mathbf{N}}^{c}\mathbf{p}_i\big\|_2^2
\end{equation}
where $\mathcal{D}_c=\{t_0, t_1, ..., t_{N_c-1}\}$ denotes the randomly sampled $N_c$ collocation points ($N_c\gg N_m$), which are used to enhance the physics satisfaction $\forall t$ (e.g., setting $N_c \geq 10N_m$ to promote the physics obeyed);  $\mathbf{N}^{m}\in\mathbb{R}^{N_m\times(r+3)}$ represents the spline basis matrix evaluated at the measured time instances while $\dot{\mathbf{N}}^{c}\in\mathbb{R}^{N_c\times(r+3)}$ is the derivative of the spline basis matrix at the collocation instances; $\boldsymbol{\Phi}\in\mathbb{R}^{N_c\times l}$ is the collocation library matrix of candidate terms. Mathematically, training the PiSL network is equivalent to solving the following optimization problem:
\begin{equation} \label{obj}
\begin{split}
    \{\mathbf{P}^*, \boldsymbol{\Lambda}^*\} &= \arg\min_{\{\mathbf{P}, \boldsymbol{\Lambda}\}} [\mathcal{L}_{d}(\mathbf{P}; \mathcal{D}_m) + \alpha \mathcal{L}_{p}(\mathbf{P}, \boldsymbol{\Lambda}; \mathcal{D}_c)] \\
    & \text{s.t.~~} \boldsymbol{\Lambda} \in \boldsymbol{\mathcal{S}}
\end{split}
\end{equation}
where $\alpha$ is the relative weighting; $\boldsymbol{\mathcal{S}}$ enforces the sparsity of $\boldsymbol{\Lambda}$. This constrained optimization problem is solved by the network training strategy discussion in Section \ref{ADO}. The synergy between spline interpolation and sparse equation discovery results in the following outcome: the splines provide accurate modeling of the system responses, their derivatives and possible candidate function terms as a basis for constructing the governing equations, while the sparsely represented equations in turn constraints the spline interpolation and project correct candidate functions, eventually turning the measured system into closed-form differential equations.

\textit{Remark: Accounting for Multiple Datasets.} When multiple independent datasets are available (e.g., due to different initial conditions (ICs)), parallel sets of cubic B-splines are used to approximate the system states corresponding to each dataset. The data loss in \eref{loss1} should be defined as the error aggregation over all datasets, while the terms used for assembling the physics loss in \eref{loss2} will be stacked since all the system responses satisfy a unified physical law.

\subsection{Network Training}\label{ADO}
The network will be trained through a three-stage strategy discussed herein, including pre-training, sparsity-promoting alternating direction optimization, and post-tuning.

\subsubsection{Pre-training}
We firstly employ a weakly physics-informed gradient-based optimization to pre-train the network. We call it ``weakly physics-informed'' because we are not given the governing equations with a concrete (sparse) form. Instead, we define an appropriate library $\boldsymbol{\phi}$ that includes all possible terms. This step simultaneously leverages the splines to interpolate the system states and generate a raw solution for non-parsimonious governing equations where the coefficients $\boldsymbol{\Lambda}$ are not constrained by sparsity. In particular, this is accomplished by a standard gradient descent optimizer like Adam or Adamax \citep{kingma2014adam} simultaneously optimizing the trainable variables $\{\mathbf{P}, \boldsymbol{\Lambda}\}$ in \eref{obj} without imposing the constraint. The outcome of pre-training will lead to a reasonable set of splines, for system state approximation, that not only well fit the data but also satisfy the general form of governing equations shown in \eref{dynamics}.

\begin{algorithm}[t!] 
\small
\SetAlgoLined
    \textbf{Input:} Library $\boldsymbol{\Phi}$, measurement $\mathcal{D}_m$, collocation points $\mathcal{D}_c$ and spline basis $\mathbf{N}$\;
    \textbf{Parameters:} $K, \delta_{tol}, M, R, \alpha, \beta, \eta$\;
    \textbf{Output:} Best solution $\tilde{\boldsymbol{\Lambda}}^\star$ and  $\mathbf{P}^\star$\;
    \textbf{Initialize:} $\boldsymbol{\Lambda}_1$, $\mathbf{P}^\star=\mathbf{P}_1$ and $\mathcal{L}^\star=\mathcal{L}(\mathbf{P}_{1}, \boldsymbol{\Lambda}_{1})$ from pre-training\; 
    \For{$k \gets 1$ \KwTo K}{
        $\boldsymbol{\Phi}=\boldsymbol{\Phi}(\mathbf{P}^\star)$\; 
        \For{$i \gets 0$ \KwTo n}{
            $\mathbf{y}_i=\dot{\mathbf{N}}^{c}\mathbf{p}^\star_{i}$\;
            $\boldsymbol{\lambda}_{i,k+1}=$  \texttt{STRidge}($\boldsymbol{\Phi}, \mathbf{y}_i, \delta_{tol}, M, R, \beta, \eta$) \;
        }
        Eliminate zeros in $\boldsymbol{\Lambda}_{k+1}$ to form $\tilde{\boldsymbol{\Lambda}}$\;
        $\big\{\mathbf{P}_{k+1}, \tilde{\boldsymbol{\Lambda}}_{k+1}\big\}=\arg\min\limits_{\{\mathbf{P}, \tilde{\boldsymbol{\Lambda}}\}}\big[\mathcal{L}_{d}(\mathbf{P}) + \alpha \mathcal{L}_{p}(\mathbf{P}, \tilde{\boldsymbol{\Lambda}})\big]$\;
        \If{$\mathcal{L}(\mathbf{P}_{k+1}, \tilde{\boldsymbol{\Lambda}}_{k+1}) < \mathcal{L}^\star$}{
          $\mathcal{L}^\star=\mathcal{L}(\mathbf{P}_{k+1}, \tilde{\boldsymbol{\Lambda}}_{k+1})$\;
          $\tilde{\boldsymbol{\Lambda}}^\star=\tilde{\boldsymbol{\Lambda}}_{k+1}$ and $\mathbf{P}^\star=\mathbf{P}_{k+1}$\;
        }
    }\caption{Hybrid ADO Strategy}\label{alg:HY}
\end{algorithm}

\subsubsection{Sparsity-Promoting Alternating Direction Optimization}
Finding the sparsity constraint subset $\boldsymbol{\mathcal{S}}$ in \eref{obj} is a notorious challenge. Hence, we turn the constrained optimization to an unconstrained form augmented by an $\ell_0$ regularizer that enforces the sparsity of $\boldsymbol{\Lambda}$. The total loss function reads:
\begin{equation} \label{l0}
    \mathcal{L}(\mathbf{P},\boldsymbol{\Lambda}) = \mathcal{L}_{d}(\mathbf{P}; \mathcal{D}_m) + \alpha \mathcal{L}_{p}(\mathbf{P}, \boldsymbol{\Lambda}; \mathcal{D}_c) + \beta\left\|\boldsymbol{\Lambda}\right\|_0 
\end{equation}
where $\beta$ denotes the regularization parameter; $\|\cdot\|_0$ represents the $\ell_0$ norm. On one hand, directly solving the optimization problem based on gradient descent is highly intractable since the $\ell_0$ regularization makes this problem NP-hard. On the other hand, relaxation of $\ell_0$ to $\ell_1$ eases the optimization but only loosely promotes the sparsity. To tackle this challenge, we develop a sparsity-promoting alternating direction optimization (ADO) strategy that hybridizes gradient decent optimization and adaptive STRidge \citep{rudy2017data}. The concept is to divide the overall optimization problem into a set of tractable sub-optimization problems, given by:
\begin{equation}
\resizebox{.89\linewidth}{!}{$
    \displaystyle
\boldsymbol{\lambda}_{i,k+1} := \arg\min_{\boldsymbol{\lambda}_i} \left[ \big\|\boldsymbol{\Phi}(\mathbf{P}_{k})\boldsymbol{\lambda}_i - \dot{\mathbf{N}}^{c}\mathbf{p}_{i,k}\big\|_2^2 + \beta{\|\boldsymbol{\lambda}_i\|}_{0} \right] \label{eq:ADO1}
$}
\end{equation}
\begin{equation}
\big\{\mathbf{P}_{k+1}, \tilde{\boldsymbol{\Lambda}}_{k+1}\big\} = \arg\min_{\{\mathbf{P},\tilde{\boldsymbol{\Lambda}}\}} \big[\mathcal{L}_{d}(\mathbf{P}) + \alpha \mathcal{L}_{p}(\mathbf{P}, \tilde{\boldsymbol{\Lambda}})\big]\label{eq:ADO2} 
\end{equation}
where $k\in\mathbb{N}$ is the alternating iteration; $\tilde{\boldsymbol{\Lambda}}$ consists of only non-zero coefficients in $\boldsymbol{\Lambda}_{k+1}=\{\boldsymbol{\lambda}_{1,k+1}, ..., \boldsymbol{\lambda}_{n,k+1}\}$; the notations of $\mathcal{D}_m$ and $\mathcal{D}_c$ are dropped for simplification. 
The hyper-parameters can be selected following the criteria: $\beta$ can be estimated by Pareto-front analysis based on pre-trained PiSL; $\eta$ is a small number (e.g., $10^{-6}$); and $\alpha$ follows the scale ratio between state and its derivative (e.g., $\alpha\sim[\sigma(\mathbf{y})/\sigma(\mathbf{\dot{y}})]^2$).
In each iteration, $\boldsymbol{\Lambda}_{k+1}$ in \eref{eq:ADO1} is determined by STRidge with adaptive hard thresholding (e.g., small values are pruned via assigning zero), while \eref{eq:ADO2} is solved by gradient descent to obtain $\mathbf{P}_{k+1}$ and $\tilde{\boldsymbol{\Lambda}}_{k+1}$ with the remaining candidate terms in $\boldsymbol{\Phi}$. Note that each column in $\boldsymbol{\Phi}$ is normalized to improve the solution posedness in ridge regression. The process is repeated for multiple ($K$) iterations until the spline interpolation and pruned equations reach a final balance (e.g., no more pruning is needed). The pseudo codes of this approach are given in Algorithms \ref{alg:HY} and \ref{STRidge}.

\begin{algorithm}[t!] 
\small
\SetAlgoLined
    \textbf{Input:} Library $\boldsymbol{\Phi}$ and data $\mathbf{y}$\;
    \textbf{Parameters:} $\delta_{tol}, M, R, \beta, \eta$\;
    \textbf{Output:} Best solution $\tilde{\boldsymbol{\lambda}}^\star$\; \vspace{1pt}
    \textbf{Baseline:} $\tilde{\boldsymbol{\lambda}}^\star=(\boldsymbol{\Phi})^{-1}\mathbf{y}, \mathcal{L}^\star = \big\|\boldsymbol{\Phi}\tilde{\boldsymbol{\lambda}}^\star-\mathbf{y}\big\|^2_2+\beta\big\|\tilde{\boldsymbol{\lambda}}^\star\big\|_0$\;\vspace{1pt}
    \textbf{Initialize:} $tol = \delta_{tol}, \boldsymbol{\Phi}_{(0)}=\boldsymbol{\Phi}$\;
    \For{$j \gets 1$ \KwTo M}{
        \For{$i \gets 1$ \KwTo R}{
            $\boldsymbol{\lambda}_{(i)}=[\boldsymbol{\Phi}_{(i-1)}^\texttt{T}\boldsymbol{\Phi}_{(i-1)}+\eta I]^{-1}\boldsymbol{\Phi}_{(i-1)}^\texttt{T}\mathbf{y}$ \;\vspace{1.5pt}
            $\tilde{\boldsymbol{\lambda}}_{(i)}=\boldsymbol{\lambda}_{(i)}[\boldsymbol{\lambda}_{(i)} \geq tol]$\;\vspace{1.5pt}
            $\boldsymbol{\Phi}_{(i)}=\boldsymbol{\Phi}_{(i-1)}[\boldsymbol{\lambda}_{(i)} \geq tol]$\;
        }
        $\mathcal{L}=\big\|\boldsymbol{\Phi}_{(R)}\tilde{\boldsymbol{\lambda}}_{(R)}-\mathbf{y}\big\|^2_2+\beta\big\|\tilde{\boldsymbol{\lambda}}_{(R)}\big\|_0$\;
        \uIf{$\mathcal{L} < \mathcal{L}^\star$}{
            $\mathcal{L}^\star=\mathcal{L}$ and $\tilde{\boldsymbol{\lambda}}^\star=\tilde{\boldsymbol{\lambda}}_{(R)}$\; 
        }\Else{
            $\delta_{tol} = \delta_{tol} / 1.618$\;
        }
        $tol = tol + \delta_{tol}$\;
    }\caption{STRidge}\label{STRidge}
    
\end{algorithm}

\subsubsection{Post-tuning}
Once we get the parsimonious form of governing equations from the above ADO process, we post-tune the control points $\mathbf{P}$ and non-zero coefficients $\tilde{\boldsymbol{\Lambda}}$ to make sure the spline interpolation and physical law are consistent. The post-tuning step is similar to pre-training except that the governing equations in post-tuning are comprised of only remaining terms and coefficients. The optimization result out of this post-tuning step is regarded as our final discovery result, where $\tilde{\boldsymbol{\Lambda}}^\star$ will be used to reconstruct the explicit form of governing equations.

\section{Numerical Experiments}\label{example}
In this section, we evaluate the efficacy of PiSL in the discovery of governing equations for two nonlinear chaotic dynamical systems based on sparsely sampled synthetic noisy data (e.g., single dataset or multi-source independent datasets, uniformly or non-uniformly sampled) and a system based on experimental data. We also compare the performance of our approach with two open source state-of-art models: Genetic-Programming-based symbolic regression (Eureqa) \citep{schmidt2009distilling} and the SINDy method (PySINDy) \citep{brunton2016discovering}. The robustness of PiSL against different levels of data noise is analyzed. The discovered equations are further validated on different datasets generated under disparate ICs to show the interpretability and generalizability. The synthetic datasets are generated by solving nonlinear differential equation by the Matlab \texttt{ode113} function. The proposed computational framework is implemented in PyTorch to leverage the power of graph-based GPU computing. All simulations in this paper are performed on a NVIDIA GeForce RTX 2080Ti GPU in a workstation with 8 Intel Core i9-9900K CPUs\footnote{Source codes/datasets are available on GitHub at \url{https://github.com/isds-neu/PiSL} upon final publication.}.

\subsection{Lorenz System}
The first example is 3-dimensional Lorenz system \citep{lorenz1963deterministic} with its dynamical behavior ($x, y, z$) governed by 
\begin{equation}\label{Eq:Lorenz} 
\begin{split}
    \dot{x}& = \sigma(y-x)\\
    \dot{y}& = x(\rho-z)-y\\
    \dot{z}& = xy-\beta z
\end{split}
\end{equation}
We consider the parameters $\sigma=10$, $\beta=8/3$ and $\rho=28$, under which the Lorenz attractor has two lobes and the system, starting from anywhere, makes cycles around one lobe before switching to the other and iterates repeatedly. The resulting system exhibits strong chaos. Gaussian white noise is added to clean signals with the noise level defined as the root-mean-square ratio between the noise and the exact solution. In particular, we consider 5\% noise in this example. 

\begin{figure}[t!]
    \centering
	\includegraphics[width=0.65\linewidth]{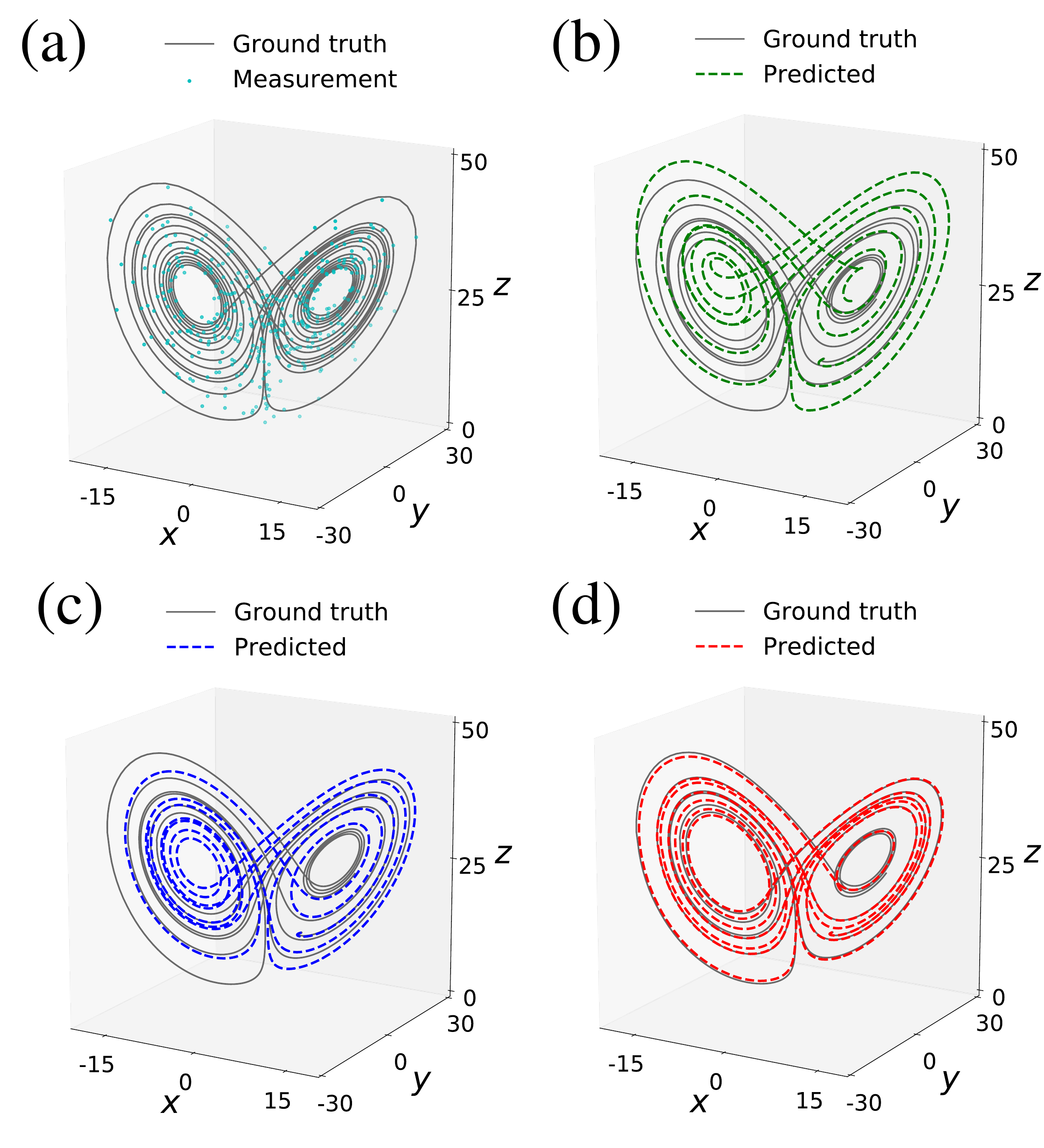} 
	\caption{The Lorenz system. (a) Sparsely sampled measurement data with 5\% noise.  Validation of (b) Eureqa-discovered equations (c) PySINDy-discovered equations and (d) PiSL-discovered equations for response prediction under different ICs.} \vspace{-6pt}
	\label{Lorenz}
\end{figure}

We first generate a single set of data uniformly sampled in [0, 20] sec at 20 Hz. Figure \ref{Lorenz}a illustrates the ground truth trajectory and the noisy measurement. We discover the equations in \eref{Eq:Lorenz} by Eureqa, PySINDy, and PiSL as well as compare their performances. The candidate function library $\boldsymbol{\phi}\in\mathbb{R}^{1\times20}$ used in PySINDy and and PiSL contains all polynomial functions of variables $\{x, y, z\}$ up to the 3rd degree. The derivatives of system state variables required by Eureqa and PySINDy are numerically approximated and smoothed by Savitzky–Golay filter. The mathematical operators allowed in Eureqa include $\{+,\times\}$, where its complexity upper bound is set to be 250. The discovered equations are listed in Table \ref{Lorenz_eqs} along with their predicted system responses (in a validation setting) shown in Figure \ref{Lorenz}. It is observed that PiSL uncovers the explicit form of equations accurately in the context of both active terms and coefficients, whereas Eureqa and PySINDy yields several false positives. The trajectory of the attractor is also much better predicted by the PiSL-discovered equations. We conclude that PiSL outperforms PySINDy and Eureqa in this discovery. On the one hand, while both PySINDy and Eureqa rely on numerical differentiation, PiSL is more robust against data scarcity and noise thanks to analytical differentiation. On the other hand, the sparsity-promoting ADO strategy is more reliable for coefficient pruning compared with the sparse identification approach in PySINDy and the insufficiency of adequate sparsity control against data noise in Eureqa, thus producing a more parsimonious solution.

\begin{table}
\centering
\caption{Discovered governing equations for the Lorenz system based on a single set of data. The blue color denotes false positives.}
{\small
\begin{tabular}{ll}
\toprule
\textbf{Model}  &  \textbf{Discovered Governing Equations} \\
\midrule
Eureqa& {\scriptsize$\dot{x} = {\color{blue} -0.56} - 9.02x + 9.01y$}\\
      & {\scriptsize$\dot{y} = {\color{blue} -0.047} + 18.79x + 1.86y {\color{blue} \,- \, 0.046xy - 0.74xz}$} \\
                & {\scriptsize$\dot{z} = {\color{blue} -3.04} - 2.23z + 0.88xy$}\\
\midrule
PySINDy& {\scriptsize$\dot{x} = {\color{blue} -0.46} - 9.18x + 9.17y$}\\
      & {\scriptsize$\dot{y} = 22.32x + 0.15y - 0.85xz$}\\
	            & {\scriptsize$\dot{z} = {\color{blue} 6.04} - 2.83z \,{\color{blue} + \, 0.15x^2} + 0.81xy$}\\
\midrule
PiSL            & {\scriptsize$\dot{x} = -10.06x + 10.03y$}\\
      & {\scriptsize$\dot{y} = 28.11x - 0.98y - 1.01xz$} \\
                & {\scriptsize$\dot{z} = -2.66z + 0.99xy$}\\
\midrule
\cellcolor{mygray} True & {\scriptsize\cellcolor{mygray}$\dot{x} = -10x + 10y$}\\
\cellcolor{mygray}      & {\scriptsize\cellcolor{mygray}$\dot{y} = 28x-y-xz$}\\
\cellcolor{mygray}      & {\scriptsize\cellcolor{mygray}$\dot{z} = -2.667z+xy$}\\
\bottomrule
\end{tabular}
}
\label{Lorenz_eqs}
\end{table}

\begin{figure}[t!]
	\centering
	\includegraphics[width=1\linewidth]{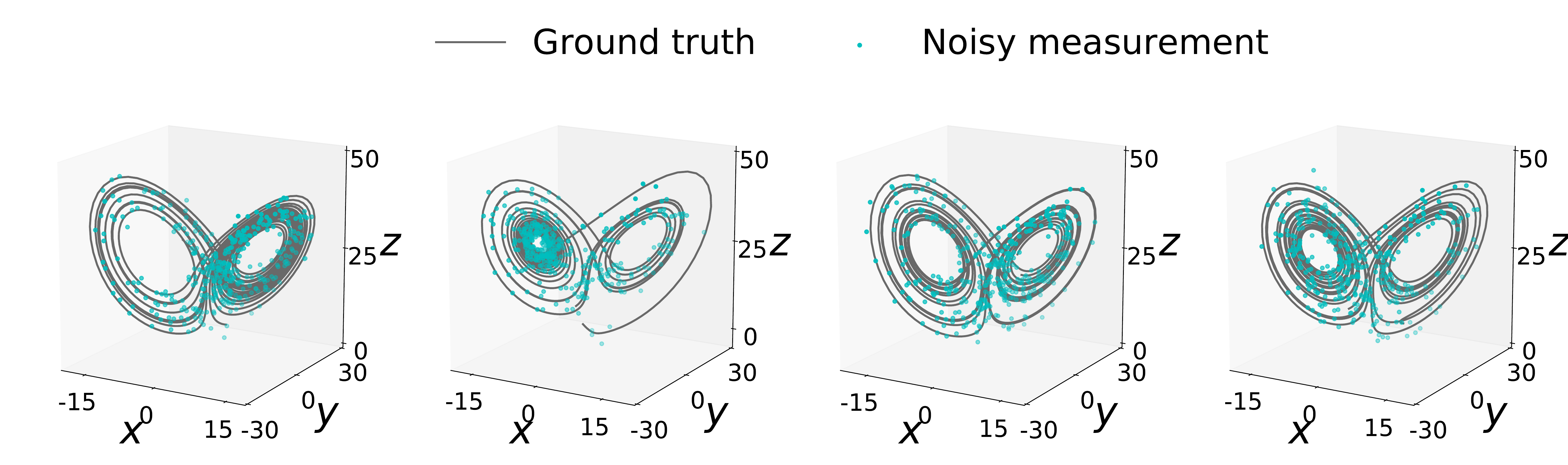} 
	\caption{Responses of four simulated Lorenz attractors for 20 seconds with different ICs. Each dataset consists of measured system states non-uniformly sampled at 200 random time points, polluted with 5\% Gaussian white noise.} \vspace{-6pt}
	\label{Lorenz multi}
\end{figure}

Next, we consider an extended case: we have multi-source measurement datasets for Lorenz attractors that are governed by the same physical law as shown in \eref{Eq:Lorenz}. Under such a circumstance, these systems, although simulated from different ICs, are in fact homogeneous. We describe their responses with a unified set of governing equations to be uncovered, which are supported by different sets of control points in splines but satisfying same physics. This ought to generate more accurate discovery since the model gathers more information on the underlying physics, despite in the presence of severe scarcity and noise of each dataset. In the following experiment, we challenge PiSL in terms of data quality by non-uniformly sub-sampling (50\% of the single dataset), which simulates the scenario of data missing or compressive sensing. We generate the noisy sub-sampled datasets under four different ICs as measurements (see Figure \ref{Lorenz multi}) for discovery. 

The discovered governing equations by PiSL based on multiple datasets are given by
\begin{equation}
\begin{split}
    \dot{x}& = -9.999x + 10.02y\\
    \dot{y}& = 27.971x - 0.999y - xz\\
    \dot{z}& = -2.666z + 0.998xy
\end{split}
\end{equation}
which are almost identical to the ground truth (see Table \ref{Lorenz_eqs}). Compared with the PiSL-discovered equations in Table \ref{Lorenz_eqs}, we can see that multi-set measurements help improve the discovery accuracy, despite the fact that the data quality and quantity of each measured attractor is worse than that in the single-set case. We conclude that, even though measurements are randomly sampled, causing the missing data issue, PiSL is still able to recapitulate the state variables through spline learning, meanwhile taking advantage of richer information from multi-set measurements to generate more accurate and less biased discovery of the governing equation for the chaotic dynamical system.

\subsection{Double Pendulum System}
\begin{figure}[b!]
	\centering
	\includegraphics[width=0.75\linewidth]{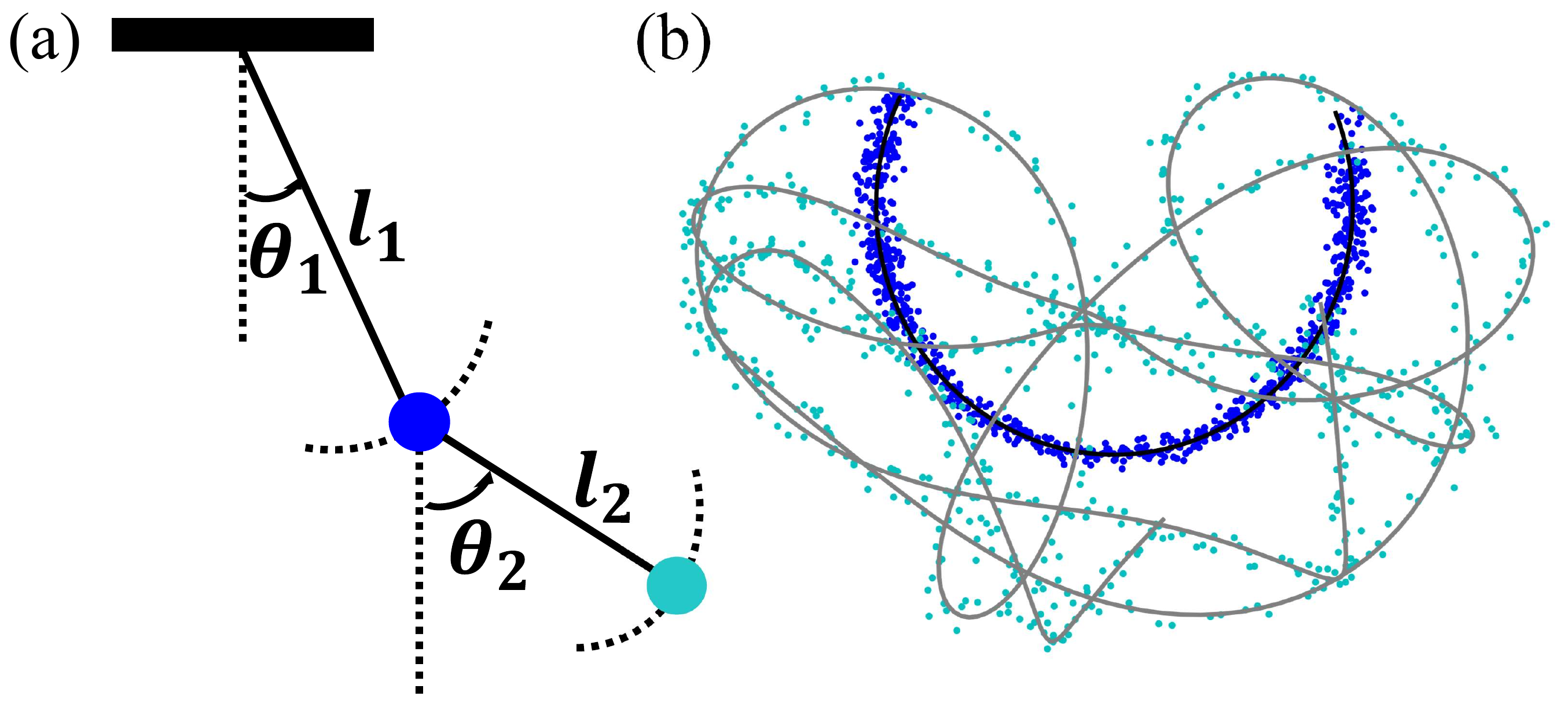} 
	\caption{The double pendulum system. (a) Schematic of the double pendulum, where the  vertically downward direction is taken as the reference origin for the angles and the counter-clockwise direction is defined as positive. (b) The measured noisy trajectories of the two pendulums, where 5\% white noise is added to the ground truth.} \vspace{0pt}
	\label{Double Pendulum}
\end{figure}

Our second numerical experiment is another chaotic system, double pendulum, as shown in Figure \ref{Double Pendulum}, which exhibits rich dynamic behavior with a strong sensitivity to ICs. In this system, one pendulum ($m_1$) is attached to an fixed end by a rod with length $l_1$, while another pendulum ($m_2$) is attached to the first one by a rod with length $l_2$. This is a classic yet challenging problem for equation discovery \citep{schmidt2009distilling,kaheman2020sindy}. The system behavior is governed by two second-order differential equations with angles $\theta_1$ and $\theta_2$ as the state variables (see Figure \ref{Double Pendulum}). These equations can be derived by the Lagrangian method:
\begin{equation} \label{DP original}
\left\{
\begin{split}
    (m_1+m_2)l_1 \dot{\omega}_1 + m_2 l_2 \dot{\omega}_2 \cos(\theta_1 - \theta_2) \, +& \\
    m_2 l_2 \omega_2^2 \sin(\theta_1-\theta_2) + (m_1+m_2)g\sin(\theta_1) &= 0, \\
    m_2 l_2 \dot{\omega}_2 + m_2 l_1 \dot{\omega}_1 \cos(\theta_1 - \theta_2) \, -& \\
    m_2 l_1 \omega_1^2 \sin(\theta_1 - \theta_2) + m_2 g \sin(\theta_2) &= 0
\end{split}
\right.
\end{equation}
where $\omega_1=\dot{\theta}_1$ and $\omega_2=\dot{\theta}_2$; $g$ denotes the gravity constant. We can see the nonlinearity from the above equations, which yield complicated behavior of the two pendulums. In fact, the behavior of this system is extremely chaotic and sensitive to the coefficients in the equations, which can be recapitulated only by the precise form of equations. We herein apply PiSL to discover the equations of motion shown in \eref{DP original} based on measured noisy trajectories.

\setlength{\tabcolsep}{0.3em}
\begin{table}[t!]
\centering
\caption{PiSL-discovered equations under different data conditions. Note that $\Delta\theta = \theta_1 - \theta_2$ and the percentage denotes noise level.}
{\small
\begin{tabular}{ll}
\toprule
\textbf{Data} &  \textbf{Discovered Governing Equations} \\
\midrule
0\%    &{\scriptsize $\dot{\omega}_1 = -0.170\dot{\omega}_2\cos(\Delta\theta) - 0.171\omega_2^2\sin(\Delta\theta) - 107.80\sin(\theta_1)$ }\\
400Hz  &{\scriptsize $\dot{\omega}_2 = -1.294\dot{\omega}_1\cos(\Delta\theta) + 1.302\omega_1^2\sin(\Delta\theta) - 139.80\sin(\theta_2)$} \\
\midrule
2\%    &{\scriptsize$\dot{\omega}_1 = -0.170\dot{\omega}_2\cos(\Delta\theta) - 0.171\omega_2^2\sin(\Delta\theta) - 107.80\sin(\theta_1)$} \\
400Hz  &{\scriptsize $\dot{\omega}_2 = -1.280\dot{\omega}_1\cos(\Delta\theta) + 1.310\omega_1^2\sin(\Delta\theta) - 138.04\sin(\theta_2)$} \\
\midrule
5\%    &{\scriptsize$\dot{\omega}_1 = -0.170\dot{\omega}_2\cos(\Delta\theta) - 0.169\omega_2^2\sin(\Delta\theta) - 107.80\sin(\theta_1)$} \\
400Hz  &{\scriptsize$\dot{\omega}_2 = -1.310\dot{\omega}_1\cos(\Delta\theta) + 1.300\omega_1^2\sin(\Delta\theta) - 140.59\sin(\theta_2)$} \\
\midrule
0\%    &{\scriptsize $\dot{\omega}_1 = -0.170\dot{\omega}_2\cos(\Delta\theta) - 0.171\omega_2^2\sin(\Delta\theta) - 107.81\sin(\theta_1)$ }\\
200Hz  &{\scriptsize $\dot{\omega}_2 = -1.263\dot{\omega}_1\cos(\Delta\theta) + 1.314\omega_1^2\sin(\Delta\theta) - 136.65\sin(\theta_2)$} \\
\midrule
\cellcolor{mygray} True & {\scriptsize\cellcolor{mygray}$\dot{\omega}_1 = -0.171\dot{\omega}_2\cos(\Delta\theta) - 0.171\omega_2^2\sin(\Delta\theta) - 107.80\sin(\theta_1)$} \\
\cellcolor{mygray}  & {\scriptsize\cellcolor{mygray}$\dot{\omega}_2 = -1.300\dot{\omega}_1\cos(\Delta\theta) + 1.300\omega_1^2\sin(\Delta\theta) - 140.14\sin(\theta_2)$} \\
\bottomrule
\end{tabular}}\label{Table:DP1}
\end{table}

\setlength{\tabcolsep}{0.5em}
\begin{table}[t!]
\centering
\caption{Data sparsity and noise effect (noise level in percentage) on PiSL and PySINDy.}
{\small
\begin{tabular}{llll}
\toprule
\textbf{Data} & \textbf{Model} & \textbf{Target Terms Found?} & \textbf{False Positives}\\
\midrule
0\%    & PiSL & Yes & 0\\
\cmidrule{2-4}
400Hz  & PySINDy & Yes & 3\\
\midrule
2\%    & PiSL & Yes & 0\\
\cmidrule{2-4}
400Hz  & PySINDy & No &NA\\
\midrule
5\%    & PiSL & Yes & 0\\
\cmidrule{2-4}
400Hz  & PySINDy & No & NA\\
\midrule
0\%    & PiSL & Yes & 0\\
\cmidrule{2-4}
200Hz  & PySINDy & Yes & 3\\
\bottomrule
\end{tabular}}\label{Table:DP2}
\end{table}

\begin{figure}[t!]
	\centering
	\includegraphics[width=0.965\linewidth]{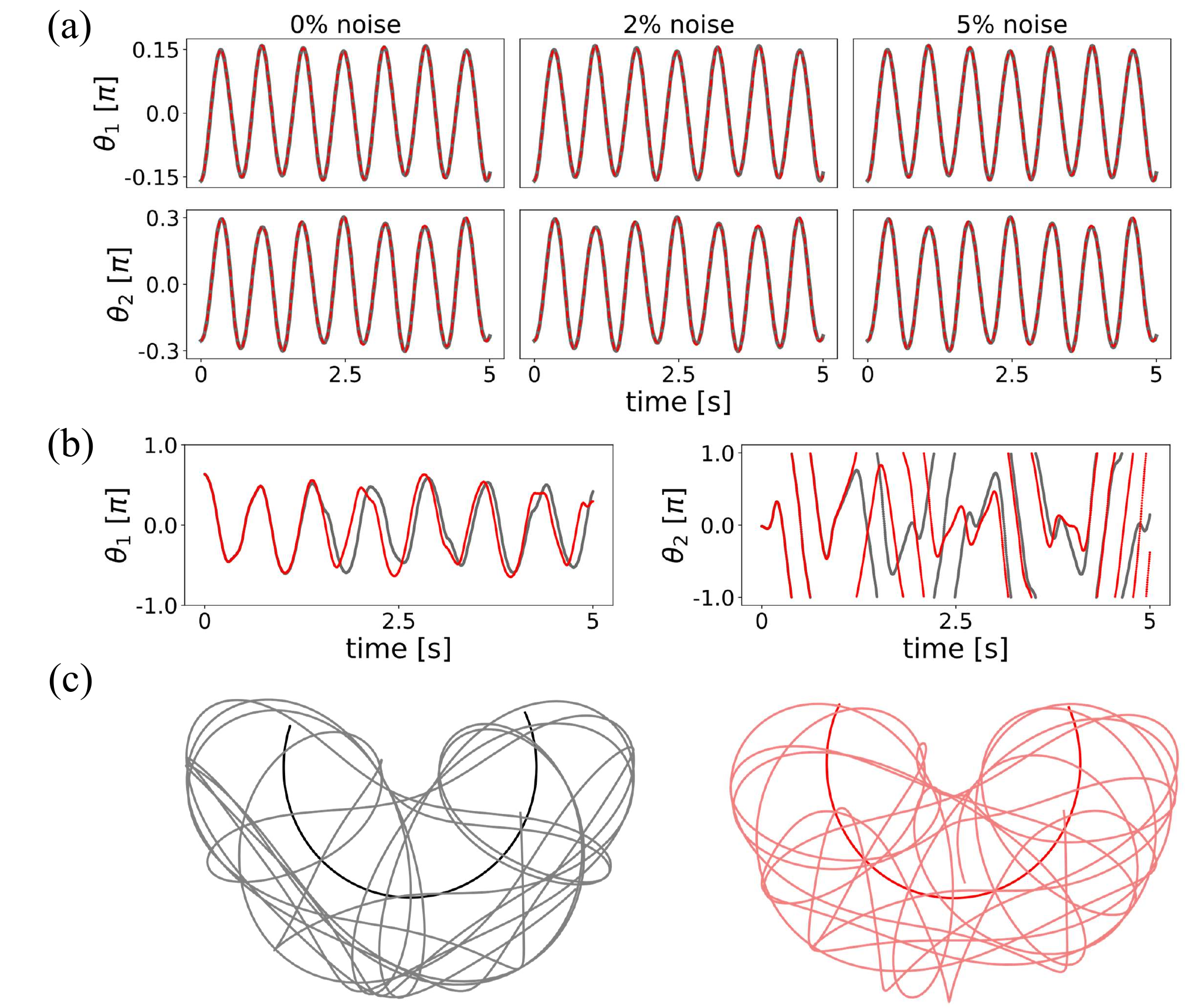} 
	\caption{Validation of the PiSL-discovered equations for the double pendulum system. (a) Predicted angular time histories within $[-\pi, \pi]$ for the case of small IC, where the red curves denote the prediction and the grey curves represent the ground truth. (b) Predicted angular time histories with $\theta_2$ exceeding $[-\pi, \pi]$, causing chaotic behavior of the second pendulum, based on 5\% noise discovery. (c) The predicted chaotic trajectories for the case described in (b).} \vspace{-6pt}
	\label{DP pred}
\end{figure}

In order to simulate a real-world situation, we consider the following parameters for a double pendulum system which match an experimental setting \citep{asseman2018learning}: $m_1=35$ g, $m_2=10$ g, $l_1=9.1$ cm, $l_2=7$ cm, with the IC of $\theta_1=1.951$ rad, $\theta_2=-0.0824$ rad, $\omega_1=-5$ rad/s and $\omega_2=-1$ rad/s. Since the angles are not directly measurable, we generate the synthetic time histories of $\{\theta_1, \theta_2\}$ and convert them to trajectories of the two pendulums, e.g., $\{x_1, y_1\}$ and $\{x_2, y_2\}$. We measure these trajectories (e.g., by video camera in practice) for 2 seconds with a sampling rate of 800 Hz and then transform back to angular time histories as measurement data for discovery. Three different noise conditions (e.g., 0\% or noise-free, 2\% and 5\%) and two subsampling frequencies (400 Hz and 200 Hz) are considered to test the robustness of PiSL and PySINDy against measurement noise and sparsity, which imitate the errors due to various precision of camera experiment setup. For this double pendulum system, we define a candidate library $\boldsymbol{\phi}$ with 20 candidate terms for both models: 
\begin{equation} \label{DP library}
\begin{split}
    \boldsymbol{\phi} =& \big\{ \phi_\theta^i\cdot\phi_\omega^j|\phi_\theta^i \in \boldsymbol{\phi}_\theta, \phi_\omega^j \in \boldsymbol{\phi}_\omega \big\} \\ &\cup
    \big\{ \dot{\omega}_1\cos(\Delta\theta), \dot{\omega}_2\cos(\Delta\theta) \big\}
\end{split}
\end{equation}
where $\Delta\theta = \theta_1 - \theta_2$, $\boldsymbol{\phi}_\theta = \{\sin(\theta_1), \sin(\theta_2), \sin(\Delta\theta)\}$ and $\boldsymbol{\phi}_\omega = \{1, \omega_1, \omega_2, \omega_1^2, \omega_2^2, \omega_1\omega_2\}$. The PiSL-discovered equations, in comparison with the ground truth for this double pendulum system, are shown in Table \ref{Table:DP1}. The effects of data noise and sampling frequency on discovery the accuracy of PiSL and PySINDy are shown in Table \ref{Table:DP2}. The closed-form expressions are successfully uncovered by PiSL with accurately identified coefficients in all data conditions considered herein, while PySINDy exhibits weaker sparsity control (e.g., yielding false positives) in the cases of noiseless datasets and evidently suffers from data noise. For validation of the PiSL-discovered equations, we consider two cases: (1) a small IC leading to periodic oscillation for both pendulums, where $\theta_1$ and $\theta_2$ never exceed the base range $[-\pi, \pi]$; (2) a large IC causing chaotic behavior of the second pendulum, where $\theta_2$ exceeds $[-\pi, \pi]$. The validation result is depicted in Figure \ref{DP pred}a for small IC and in Figure \ref{DP pred}b-c for large IC. It is seen that, although the small oscillations can be well predicted, the chaotic behavior is intractable to recapitulate (only the dynamics within the first second is accurately captured). This is due to the fact that, for ICs which are large enough to cause chaos, the response is extremely sensitive to the equation coefficients, even in the presence of a tiny difference.

\subsection{EMPS}
\begin{figure}[t!]
	\centering
	\includegraphics[width=0.98\linewidth]{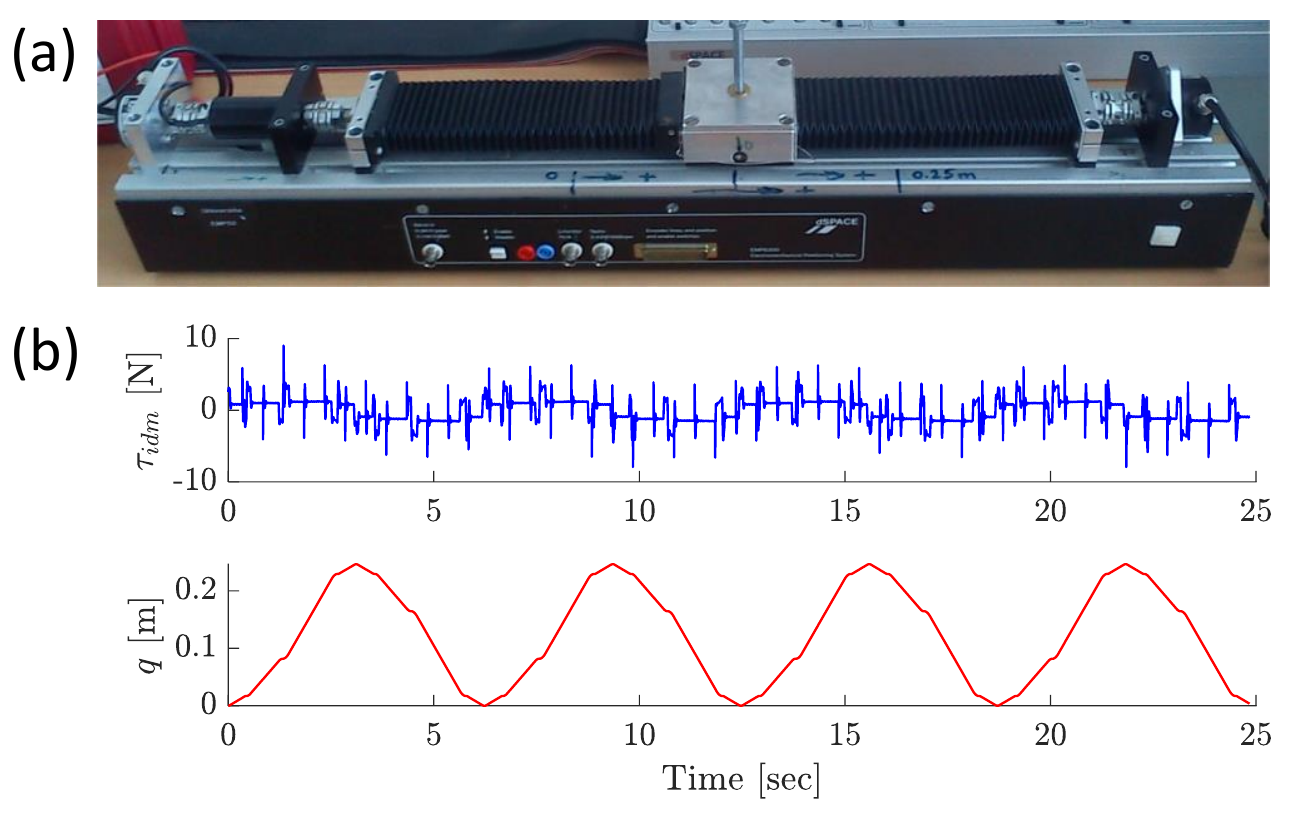} 
	\caption{\small The Electro-Mechanical Positioning System: (a) device setup, (b) measured input and output data.} 	
	\label{Figure:EMPS1}
\end{figure}

An experimental example is finally considered in this case, aka., an Electro-Mechanical Positioning System (EMPS) \citep{janot2019data}, shown in Fig. \ref{Figure:EMPS1}a, which is a standard configuration of a drive system for prismatic joint of robots or machine tools. The main source of nonlinearity is caused by the friction-dominated energy dissipation mechanism. A possible continuous-time model suitable to describe the forced vibration of this nonlinear dynamical system is given by:
\begin{equation} \label{eq:EMPS}
    \ddot{q} = \frac{1}{M}\tau_{idm}-\frac{F_v}{M}\dot{q}-\frac{F_c}{M}\mathrm{sign}(\dot{q})-\frac{1}{M}c
\end{equation}
where $q$ (the output), $\dot{q}$ and $\ddot{q}$ are the joint position, velocity and acceleration, respectively; $\tau_{idm}$ is the joint torque/force (the input); $c$ is a constant offset; $M$, $F_v$ and $F_c$ are referred to as constant parameters. We introduce a variable $p=\dot{q}$ denoting velocity to convert \eref{eq:EMPS} to the target form of first-order differential equation in a state-space formulation. Fig. \ref{Figure:EMPS1}b shows the measurement data.

We define the candidate library as $\{q, q^2, p, p^2, \mathrm{sign}(p), \tau,$ $1\}$ for PiSL and PySINDy, while allowing $\{+, \times, \mathrm{sign}\}$ and constant as the candidate operations in genetic expansion and set upper bound of complexity to be 50 in the Eureqa approach. The discovery results are reported in Table \ref{Table:EMPS}, in comparison with a reference target equation where the values of coefficients are found in \citet{janot2019data}. We observe that PiSL and Eureqa produce the same equation form with close parameter estimation while PySINDy fails to enforce sparsity in the distilled equation. Given the sparse data, the $p^2$ term in the discovered equations by PiSL and Eureqa supersedes the small offset constant that has a minor effect.

\setlength{\tabcolsep}{0.3em}
\begin{table}[t!]
\centering
\caption{Discovered governing equations for the EMPS system. }
{\small
\begin{tabular}{ll}
\toprule
\textbf{Model}  &  \textbf{Discovered Governing Equations} \\
\midrule
Eureqa       & {\scriptsize$\dot{p}=0.368\tau-2.248p-0.202\mathrm{sign}(p) + 0.0141 + 2.453p^2$} \\
\midrule
PySINDy      & {\scriptsize$\dot{p}=0.368\tau-0.112p-0.212\mathrm{sign}(p) +0.0195 - 2.144q$}\\
 &  {\scriptsize $~~~~~~~+ 0.294p^2 + 2.547q^2$} \\
\midrule
PiSL           & {\scriptsize$\dot{p}=0.369\tau-2.276p-0.20\mathrm{sign}(p) + 0.0121 + 2.547p^2$} \\
\midrule
\cellcolor{mygray}Reference   & {\scriptsize\cellcolor{mygray}$\dot{p} = 0.370\tau+2.140p+0.214\mathrm{sign}(p) + 0.0333$} \\
\bottomrule
\end{tabular}
}
\label{Table:EMPS}
\end{table}

\section{Conclusion}
In this paper, we propose a physics-informed spline learning method to tackle the challenge of distilling analytical form of governing equations for nonlinear dynamics from very limited and noisy measurement data. This method takes advantage of spline interpolation to locally sketch the system responses and performs analytical differentiation to feed the physical law discovery in a synergistic manner. Beyond this point, we define the library of candidate terms for sparse representation of the governing equations and use the physics residual in turn to inform the spline learning. We must acknowledge that our approach also has some limitations, e.g, (1) limited capacity of linear combination of candidate functions to represent very complex equations, (2) incorrect discovery given improperly designed library, and (3) inapplicable to problems where system states are incompletely measured or unmeasured. Despite these limitations, 
the synergy between splines and sparsity-promoting physics discovery leads to a robust capacity for handling scarce and noisy data. Numerical experiments have been conducted to evaluate our method on two classic chaotic systems with synthetic datasets and one system with experimental dataset. Our proposed model outperforms the state-of-the-art methods by a noticeable margin and conveys great efficacy under various high-level data scarcity and noise situations.


\clearpage

\bibliographystyle{named}
\bibliography{ijcai21}

\end{document}